\pdfoutput=1

\documentclass[11pt,UTF8]{article}

\usepackage[]{emnlp2021}

\usepackage{times}
\usepackage{latexsym}

\usepackage[T1]{fontenc}

\usepackage[utf8]{inputenc}

\usepackage{microtype}

\usepackage{bm}
\usepackage{amsmath}
\usepackage{booktabs}
\usepackage{float}
\usepackage{amsmath}
\usepackage{graphicx}
\usepackage{CJKutf8}

\newcommand{\red}[1]{\textcolor{red}{#1}}
\newcommand{\blue}[1]{\textcolor{blue}{#1}}

\title{Extract, Integrate, Compete: \protect\\Towards Verification Style Reading Comprehension}

\author{
    Chen Zhang, 
    Yuxuan Lai, 
    Yansong Feng\thanks{\;\;Corresponding author.}~~,  
    Dongyan Zhao \\
    Wangxuan Institute of Computer Technology, Peking University, China\\
    The MOE Key Laboratory of Computational Linguistics, Peking University, China\\
    {\tt \{zhangch, erutan, fengyansong, zhaody\}}
    {\tt @pku.edu.cn} \\
}

\begin{document}
\begin{CJK}{UTF8}{gbsn}
\maketitle
\begin{abstract}
In this paper, we present a new verification style reading comprehension dataset named VGaokao from Chinese Language tests of Gaokao. Different from existing efforts, the new dataset is originally designed for native speakers' evaluation, thus requiring more advanced language understanding skills. To address the challenges in VGaokao, we propose a novel Extract-Integrate-Compete approach, which
 iteratively selects complementary evidence with a novel query updating mechanism and adaptively distills supportive evidence, followed by a pairwise competition to push models to learn the subtle difference among similar text pieces.
Experiments show that our methods outperform various baselines on VGaokao with retrieved complementary evidence, while having the merits of efficiency and explainability. Our dataset and code are  released for further research\footnote{\url{https://github.com/luciusssss/VGaokao}}.

\end{abstract}

\section{Introduction}
Reading comprehension has been frequently used in various standardized examinations to evaluate one's language understanding skills, where test-takers are expected to read a long article, answer a series of questions, or verify a given statement according to the passage.   
For example, in the Chinese Language tests of Gaokao (
also known as 
China National College Entrance Examination), approximately half of the reading comprehension questions are in a \textit{verification} style. 
As shown in Table~\ref{tab:c3} (bottom), students are expected to read a passage, and then select from four choices (A\textasciitilde D) the best statement that is the most consistent with the passage, or sometimes, contracts the most to the passage.

While the question answering styled tasks have been intensively studied in the {NLP community~\cite{squad,lai2017race,hotpot,sun2020investigating}}, the verification styled MRC task actually receives much less attention.   
Here, as indicated in the Gaokao Instructions\footnote{\url{http://gaokao.neea.edu.cn/}}, the abilities of gathering multiple evidence pieces from long articles, distilling supportive evidence, and  making decisions accordingly by capturing the subtle
difference among similar text pieces (i.e., choices), are necessary skills for Chinese Language learning.
This type of questions actually provides an ideal test-bed for natural language understanding research.

\begin{figure}[t]
\centering
\includegraphics[width=1\linewidth]{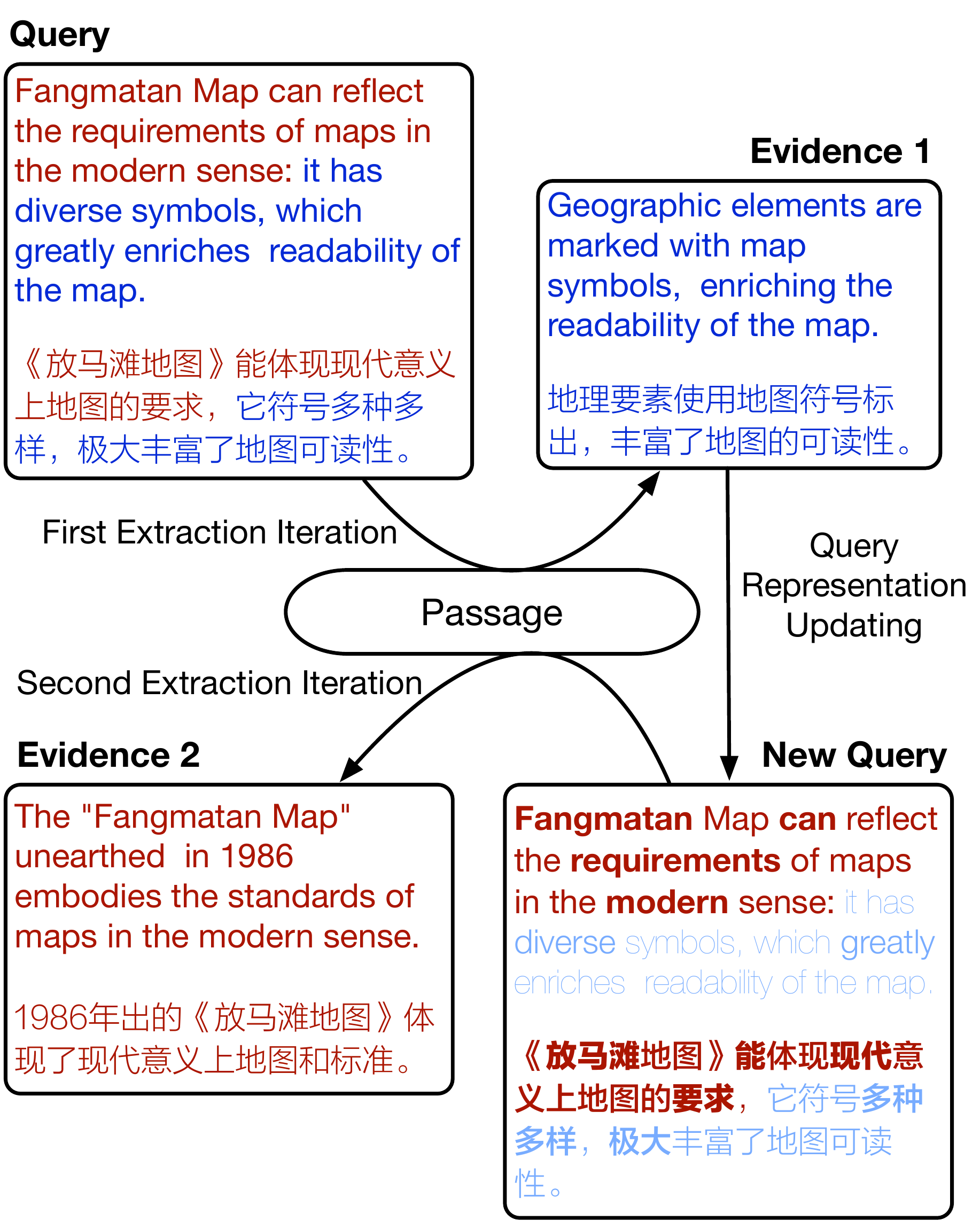}
\caption{An example of 
iterative evidence extraction. 
The darker tokens in \textbf{bold}  
 are more important 
for the 
updated query.}
\label{fig:example}
\end{figure}

In this paper, we present a verification style reading comprehension dataset named \textbf{VGaokao} to highlight the language understanding challenges 
mentioned above.
VGaokao is constructed from the Chinese Language tests of Gaokao.
Figure~\ref{fig:example} shows an example in VGaokao, where a statement (one of the 4 choices) should be verified according to the given passage.
To do so, we need to
extract 
two evidence sentences from the passage, combine them together to examine how well the statement is supported by the evidence, and finally compare with other choices to arrive at the answer.

As constructed from standardized language test for native speakers, VGaokao involves more language understanding challenges than the datasets constructed from tests for second language learners such as C$^3$~\cite{sun2020investigating} and RACE~\cite{lai2017race}. 
For example, VGaokao involves more diverse vocabulary and more complex sentence structures. Besides, nearly half of the statements in VGaokao require multiple evidence to verify.
Compared with the fact checking tasks such as FEVER~\cite{thorne2018fever}, where each claim is 
tagged with a definite label individually, most statements in VGaokao are not absolutely correct nor absolutely incorrect, which 
requires models to carefully compare 
one statement with another to choose the most suitable answer according to the given passage. 

To address the above challenges in VGaokao, we propose a novel  
 \textbf{Extract-Integrate-Compete} framework, 
where two query updating strategies, hard masking and soft masking, are designed to iteratively extract multiple complementary evidence for a given statement.
Figure~\ref{fig:example} shows an example of using soft masking to highlight the tokens whose corresponding evidence has not been found up to current iteration. 
After evidence extraction, we adaptively filter irrelevant evidence sentences and dynamically determine the number of evidence pieces to be integrated. 
Options in each question are then verified against retrieved evidence and are compared in a pairwise style to select the most plausible answer.

We empirically investigate the performance of our Extract-Integrate-Compete approach on VGaokao. 
Experiments shows that our method outperforms end-to-end methods with passage chunking and various evidence selection methods in evidence retrieval F1. Performance gains in evidence selection can further propagate to the final question answering performance.

Our contributions can be summarized as: 1) We propose a novel verification style reading comprehension dataset, VGaokao, which embeds more advanced language understanding skills. 
2) We propose a new Extract-Integrate-Compete approach to iteratively select complementary evidence from long articles through a novel query updating mechanism. Our  hinge loss based competition component can push the model to capture fine-grained differences among different choices. 3) Experiments show that our approach outperforms a variety of baselines in both evidence retrieval F1 and QA accuracy on VGaokao while showing the merits of efficiency and explainability.

\section{VGaokao: Verification Style Reading Comprehension Dataset}
\label{sec:dataset}
Standardized language tests have been considered as a test-bed to harvest machine reading comprehension datasets. While most existing efforts focus on SQuAD-like QA datasets \cite{hotpot,kwiatkowski2019natural}, or cloze style  questions \cite{record,zheng2019chid}, few efforts are made to verification style questions.

In the Chinese Language test of Gaokao, approximately half of the reading comprehension questions instruct students to select a statement (i.e., an option from four choices) that is the most consistent or contradicting with the given passage.
These questions are designed to evaluate students' ability in \textit{extracting and integrating information from long passages, and analyzing certain linguistic phenomena or semantic relations among several similar sentences}\footnote{According to the Syllabus of Chinese Language in China by the Ministry of Education.}.
According to the target language skills, we call these questions \textit{verification style questions} and convert them into the task of verifying given statements against the articles. Our task is similar in spirit to the fact checking task, FEVER~\cite{thorne2018fever}. But, unlike the claims tagged with a definite label in FEVER (\textit{Supported}, \textit{Refuted} or \textit{Not-Enough-Info}), the statements in Gaokao questions are designed to be not absolutely correct nor absolutely incorrect. 
In other words, the options of a question in VGaokao differ in how much they are supported by the passage, which is difficult to be quantified with a value between 0 and 1.
The design of relative correctness is in the purpose of evaluating test-takers' ability to disambiguate very similar options through considering the subtle difference among them. 

In this work, we construct a verification style reading comprehension dataset named VGaokao from the Chinese Language test of Gaokao to assess a model's skills of extracting evidence from articles and verifying statements against the retrieved evidence pieces and other options.

\begin{table*}[t]
\centering
\small
\setlength{\tabcolsep}{8pt}
\begin{tabular}{p{0.35\textwidth}p{0.55\textwidth}}

\toprule 
{\normalsize \textbf{C${^3}$}} &  \\
{\small ...他呆呆地站在那里，面色尴尬至极，双手拧来拧去无处可放。上课前他自以为成竹在胸，所以就没带教案和教材。整整10分钟，教室里鸦雀无声，所有的学生都好奇地等着这位新来的老师开口...{(共498字)}}

{\small \textbf{Q:}~沈从文没拿教材，是因为他觉得（ ）}

{\small A. 讲课内容不多}

{\small B. 自己准备得很充分 $\surd$}

{\small C. 这样可以减轻压力}

{\small D. 教材会限制自己的发挥}
&
{\small ...He stood there motionlessly, extremely embarrassed. He wrung his hands without knowing where to put them. Before class, he believed that he had had a ready plan to handle different situation so he did not bring his teaching plan and textbook. For up to 10 minutes, the classroom was in perfect silence. All the students were curiously waiting for the new teacher to open his mouth...
{(498 Chinese characters in total)}}

{\small \textbf{Q:}~Congwen Shen did not bring the textbook because he felt that ( ) }

{\small A. there were not many teaching contents.}

{\small B. he was well prepared. $\surd$}

{\small C. he could relieve his mental pressure in this way.}

{\small D. the textbook was likely to restrict his ability to give a lecture.}\\
\midrule
{\normalsize \textbf{VGaokao}} & \\
{\small ...如果孕妇怀孕期间睾丸素水平较高，生下的孩子就更容易成为左撇子，也更容易在日后得心脏疾病或孤僻症。这也许可以解释为什么世界上左撇子只占总人口的10％。同样，雌性激素水平较高的孕妇，所生的女孩食指通常短于无名指，而日后患乳腺癌的可能性也较高。... {(共1126字)}}

{\small {\textbf{Q:}} 下列说法符合原文意思的一项是（ ）} 

{\small A. 在400万年进化史中，人类的手逐渐演变成使人具有高度智慧的重要器官和大自然所能创造出的最完美的工具。}

{\small B.通常，左撇子的无名指之所以比食指长，是因为孕妇怀孕期间的睾丸素水平较高。$\surd$}

{\small C.大脑控制手的活动的区域的运动中枢与语言中枢之间存在着密切的神经联系，使得手势成为人类沿袭至今的唯一的肢体语言。}

{\small D. 大脑控制手的活动区域，面积达大脑皮层1/4。因此大脑皮层特别强烈的兴奋会使一个简单的手动作顺利实现。}
&
{
\small ...If pregnant women have higher testosterone levels during pregnancy, their children are more likely to be left-handed and more likely to develop heart disease or autism in the future. This may explain why left-handers only account for 10\% of the total population in the world. Similarly, pregnant women with higher levels of estrogen usually give birth to girls whose index finger is shorter than the ring finger, and they are more likely to develop breast cancer in the future... {(1126 Chinese characters in total)}
}

{\small {\textbf{Q:}} Which statement is most consistent with the passage? ( ) }

{\small A. In 4-million-year evolutionary history, human hands have gradually evolved into an essential organ enabling human to have high-degree intelligence and the most perfect naturally created tool.}

{\small B. In general, the fact that the left-handed ring finger is longer than the index finger is because of the higher testosterone levels during pregnancy.$\surd$}

{\small C. There is a close neural connection between the motor center and the language center in the area where the brain controls the activity of the hands, making gestures the only body language that has been inherited by humans.}

{\small D. The brain controls the active area of hands, which reaches 1/4 of the cerebral cortex. Thus, particularly intense excitement in cerebral cortex makes a simple hand movement smoothly realized.} \\

\bottomrule
\end{tabular}
\caption{Example articles and questions from C$^3$~\cite[top]{sun2020investigating} and from VGaokao (bottom).}
\label{tab:c3}
\end{table*}

\begin{table}
\centering
\setlength{\tabcolsep}{6pt}
\small
\begin{tabular}{l|rrr}
\toprule
        & Passage & Question & Option \\
\midrule
Average Length (Char.)  & 1,159 & 18 & 47\\
Max Length (Char.) & 2,568 & 60 & 147  \\
Vocabulary Size & 96,806 & 1,245 & 34,789 \\
\midrule
Total Vocabulary Size & \multicolumn{3}{c}{87,945} \\
\bottomrule
\end{tabular}
\caption{Statistics of the VGaokao dataset.}
\label{tab:stats}
\end{table}

\subsection{Dataset Construction}
We construct VGaokao from the Chinese Language examinations in Gaokao and the official mock tests provided by each province.
The test questions are designed by the Ministry of Education of China and the examination centers of each province, which are available to the public. 
We first collect the original set of test article-question pairs and discard 
the articles and questions that are not of verification style\footnote{We filter out fictions, proses, and poems from the collection, which focus on evaluating students' 
ability of aesthetic appreciation.
We also discard the questions asking for word meaning explanation or summarizing a certain topic.
}. 
The remaining articles cover a wide range of topics, including analyzing issues or arguments, introducing recent discoveries in science or engineering, and discussing popular social topics. The remaining questions ask the test-takers to select a statement that is the most consistent with the article or contracts the most with the article, which we can convert into the form of statement verification. 

In total, we collect 2,786 passages and 3,512 questions.
Each question is paired with 4 options.
We randomly sample 80\% of the questions for training and the rest of the questions are used for test.

To quantitatively evaluate the model performance in evidence selection, we randomly sample 25 questions (100 options) from the test set of VGaokao and manually annotate their evidence sentences from the article.
On average, there are {1.6} evidence sentences for each option, and the distance between evidence varies {from 1 to 9} sentences.
This indicates that one has to collect 6.4 evidence on average from the whole passage to answer one question with four options, which makes VGaokao a dataset requiring the ability of gathering and processing multiple evidence.

\subsection{Dataset Analysis}
The basic statistics of VGaokao is shown in 
Table~\ref{tab:stats}, illustrating
challenges from the following aspects.

\paragraph{Advanced Language Comprehension} 
Constructed from standardized language tests designed for native speakers, VGaokao requires models to understand more complex passages, compared to other datasets from examinations for second language learners such as C$^3$~\cite{sun2020investigating}.
The main differences can be summarized in three folds.
Firstly, in VGaokao, the vocabulary size is 1.6 times larger than that of C$^3$, which brings more diverse words.
Secondly, at the sentence level, the average sentence length of VGaokao is 1.6 times longer than that of C$^3$, and the average dependency tree depth of VGaokao is 1.2 times larger than that of C$^3$. 
Longer sentences and more complicated syntactic structures usually exhibit rich linguistic phenomena, thus requiring models to learn more sophisticated language understanding skills.
Lastly, the passages in VGaokao, which contain 1,159 Chinese characters on average, are approximately 3 times longer than that of RACE~\cite{lai2017race} and 10 times longer than that of C$^3$~\cite{sun2020investigating}.
The length of most passages even exceeds the maximum input length of general pre-trained language models such as BERT\cite{bert}. 
To exploit the long passages, models may need to take discourse structures into consideration so as to better integrate multiple evidence sentences.

As shown in  the sample articles and questions from VGaokao and C$^3$ in Table~\ref{tab:c3}, the example article from VGaokao involves domain-specific terminologies such as \textit{testosterone} and \textit{estrogen}. 
The sentences in the VGaokao example are longer and involves more complicated sentence structures such as compound sentences. Besides, the options in VGaokao seem to be more confusing: Option C and D both discuss the subtle relationship between the brains and hands, which requires one to carefully discriminate between them based on the passage, while in the C$^3$ example, only Option B is explicitly mentioned in the passage.
These comparisons indicate that VGaokao can be used to evaluate the language comprehension ability at a higher level compared to previous works.

\paragraph{Requirement of Multiple Evidence}
Longer passages usually require skills of gathering multiple disjoint evidence pieces to verify and compare the candidate statements.
As a pilot study, we randomly sample 100 options and manually annotate their golden evidence sentences from their corresponding passages.
We find that 47\% of the sampled options require more than one evidence sentences. 
Furthermore, there are usually certain discourse relationship among multiple evidence pieces required by one statement, such as causality, comparison, or expansion. 
In the VGaokao example of Table~\ref{tab:c3}, to support Option B, we need to extract the first and the third sentences in the sample text as evidence and recognize the causal relationship between them. 
This observation indicates that more sophisticated skills such as discourse analysis may be helpful for exploiting the evidence pieces.

\begin{figure*}[t]
\centering
\includegraphics[width=1.0\linewidth]{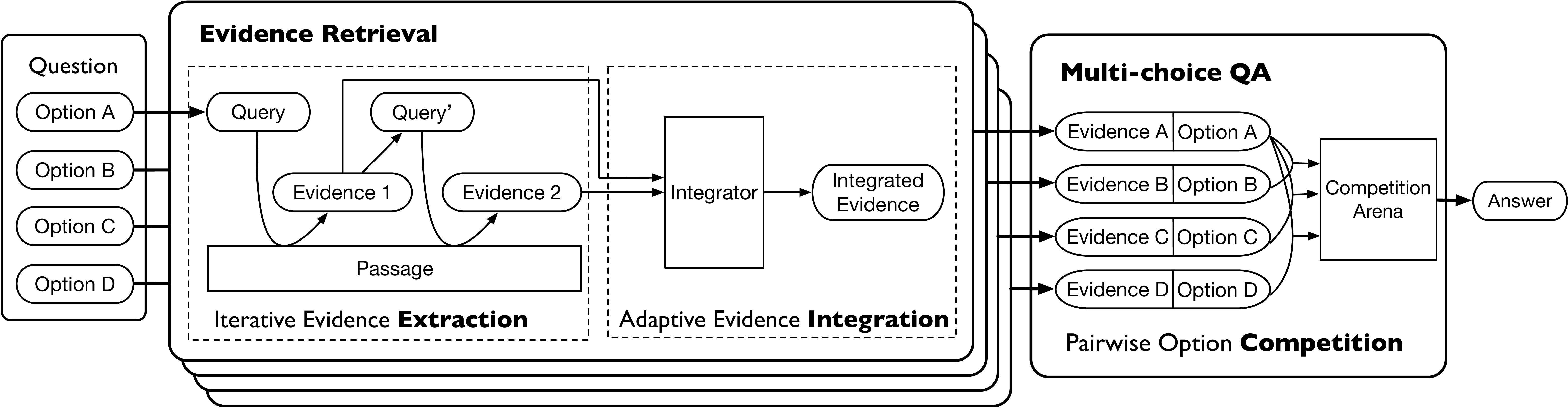}
\caption{An illustration of our proposed Extract-Integrate-Compete approach.}
\label{fig:pipeline}

\end{figure*}

\section{Our Extract-Integrate-Compete Approach}

We propose a novel Extract-Integrate-Compete approach to address the challenges in VGaokao.  As illustrated in Figure~\ref{fig:pipeline}, our approach includes three stages, iterative evidence extraction, adaptive evidence integration and pairwise option competition. 

\subsection{Iterative Evidence Extraction}

Generally, in evidence sentence extraction, models are to extract a subset of evidence sentences $\left\{s_1,~s_2,~\cdots,~s_n\right\}$ according to query $q$.
One can use an encoder (e.g., averaging over the presentation of each word in the embedding space) to embed query $q$ and evidence candidate $s_i$ into dense vectors $\bm{q}$ and $\bm{s_i}$, and then use a similarity function $\rm{sim}(\cdot)$ (e.g., cosine similarity) to obtain the relevance score $f$,
i.e., $f(q, s_i) = {\rm{sim}}(\bm{q}, \bm{s_i})$.
This method treats each candidate evidence independently, and
produces a ranking list. 
However, when a query requires multiple evidence sentences, 
independently
selecting top $k$ ranked sentences may ignore the complementary 
relationship between evidence sentences, thus producing inferior results.

In this work, we propose an iterative way to collect multiple evidence for a given query,
where the query representation is updated with newly-retrieved information at each extraction iteration.
To emphasize the complementary relation between the evidence sentences while avoiding too much 
overlap, we propose to use \textbf{masking strategy} to reduce the relevance between queries and their retrieved evidence pieces.

Specifically, 
we assume 
$e^t=\{c^t_1,~c^t_2,~$$\cdots,~c^t_m\}$ is the evidence sentence retrieved at the the $t$-th step, $c^t_j$ is the $j$-th token in $e^t$.
To search for complementary evidence pieces, we propose to
reduce the influence of some parts in $q$ that have been covered by the retrieved evidence sentence $e_t$
so that the query representation $\bm{q_{t+1}}$ for the next step will pay more attention to the parts whose corresponding evidence has not been found.

We design two masking strategies for iterative evidence extraction, i.e., \textbf{Hard Masking} and \textbf{Soft Masking}.

\textbf{Hard Masking}\quad 
After the $t$-th iteration, we just discard the query tokens that exactly appear in the extracted evidence $e^t$. 
Formally, in the query representation $\bm{q_{t+1}}$ for $(t+1)$-th step, the weight $\beta^t_{i}$ of the $i$-th query token $q_i$ is:
\begin{equation}
\beta^{t+1}_{i} = \frac{\alpha^{t+1}_i}{\sum_k  \alpha^{t+1}_k}
\end{equation}

\begin{equation}
\alpha^{t+1}_i = \left\{
\begin{array}{rr}
0,       &  q_i \in {e^t}  \\
\alpha^{t}_i,       &  q_i \notin {e^t}
\end{array} \right.
\end{equation}

Afterwards, we compose the new query representation $\bm{q_{t+1}}$ for the $(t+1)$ step by performing a weighted sum of the query token embeddings:
\begin{equation}
\bm{q^{t+1}} = \sum_{i} \beta^{t+1}_{i} \bm{q_i}
\label{eq4}
\end{equation}

In this way, the new query representation is restricted to focus on the unmatched tokens.

\textbf{Soft Masking}\quad
Instead of directly assigning zero weights in the hard masking method, soft masking strategy 
reduces
the weights of the already-addressed tokens in the next-step representation. 
The weight of the $i$-th query token $q_i$ is inversely correlated to its matching score to its most similar token in the retrieved evidence set. In practice, we use the dot products of token embeddings to measure the relevance between tokens. Then we calculate the weight $\beta^{t+1}_{i}$ for the $i$-th query token at the $(t+1)$-th retrieval step by adding a negative sign to its highest similarity score and applying a softmax over all the query token weights:
\begin{equation}
\beta^{t+1}_{i} = \frac{e^{\lambda \alpha^{t+1}_i}}{\sum_k e^{\lambda \alpha^{t+1}_k}}
\label{eq:lambda}
\end{equation}
\begin{equation}
\alpha^{t+1}_i = -\max{\left(\max_{j} (\bm{q_i} \cdot \bm{c^t_j}),-\alpha^{t}_i\right)}
\end{equation}
where $\lambda$ is used to adjust the extent to which we want to widen the weight gap between the matched tokens and the unmatched tokens.

Afterwards, we can obtain the new query representation $\bm{q_{t+1}}$ similarly to Eq.\ref{eq4}.

\subsection{{Adaptive Evidence Integration}}
In practice, different queries require various amount of evidence pieces. 
Fixing the numbers of evidence sentences
may introduce noise 
for queries requiring fewer or more evidence sentences. 
To alleviate this problem, we introduce an evidence integration module
to adaptively determine how many complementary evidence pieces are needed for each query.

Specifically, after $t$ steps of beam search, we obtain several evidence chains consisting of $t$ different evidence sentences.
In each chain, we reorder and concatenate the evidence sentences according to their order in the original passages, in the hope to maintain the potential discourse relationship between evidence sentences. 
Then, we feed the integrated evidence chains obtained from different numbers of retrieval steps into a reranker, further comparing their semantic similarity to the query. The highest scored evidence chain will be selected as the final evidence chain of the query.
This evidence integrator measures the candidate evidence chains as a whole and adaptively filters irrelevant evidence pieces introduced in later iterations.

\subsection{{Pairwise Option Competition}}
As mentioned in Section~\ref{sec:dataset}, 
we need to carefully discriminate among several options to arrive at the final answer. We thus introduce a pairwise option competition component
to the option selection step.
We adopt a pre-trained language model $g(\cdot)$ to calculate how the statement $d$ is supported by a retrieved evidence set $c$.  
During training, for a question that requires choosing the option that is most consistent with the passage,  we have a correct option $d^{+}$ and several incorrect options $d^{-}_1, d^{-}_2, ..., d^{-}_k$, paired with their retrieved evidence sets $c^{+}, c^{-}_1, c^{-}_2, ..., c^{-}_k$. 
We calculate a hinge loss for the pairwise option competition:
\begin{equation}
    \begin{aligned}
    & L(d^{+}, d^{-}_1, d^{-}_2, ..., d^{-}_k) = \\
    &  \sum_{i=1}^{k} \max (0,  -g(d^{+}, c^{+}) + g(d^{-}_i, c^{-}_i) + 0.5)
    \end{aligned}
\end{equation}

Similarly, for a question that requires choosing the most contradictive option, we will have an incorrect option $d^{-}$ and several correct options $d^{+}_1, d^{+}_2, ..., d^{+}_k$, paired with their retrieved evidence chains $c^{-}, c^{+}_1, c^{+}_2, ..., c^{+}_k$, the loss will be:
\begin{equation}
    \begin{aligned}
    & L(d^{-}, d^{+}_1, d^{+}_2, ..., d^{+}_k) = \\
    &  \sum_{i=1}^{k} \max (0,  - g(d^{+}_i, c^{+}_i) + g(d^{-}, c^{-}) + 0.5)
    \end{aligned}
\end{equation}

During inference, we select the option with the highest score as the answer for a question requiring choosing the option that is the most consistent to the passage.
Similarly, we select the option with the lowest score for a question asking for the option that contradicts most to the passage.

\subsection{Implementation Details} 
For the iterative extractor, we use jieba\footnote{\url{https://github.com/fxsjy/jieba}} to perform Chinese word segmentation.
We use pre-trained word vectors \cite{qiu2018revisiting} to perform unsupervised iterative extraction with our query updating strategies.  The $\lambda$ in Eq.~\ref{eq:lambda} is set to 1. 
Since 83\% of the options with multiple evidence pieces require two evidence sentences, we set the maximum number of iterations to 2. 
Considering that maintaining all evidence chains during iterative extraction has exponential complexity to the steps, we use beam search, where only top 2 evidence sentences remain in each step.

We use Sentence-BERT~\cite{reimers2019sentence,reimers2020making} to measure the relevance between the query and the evidence chains in the adaptive integrator.
For the pairwise option competition, we use 
Chinese RoBERTa-wwm-ext-Large~\cite{cui2019pre} with Transformers toolkit~\cite{wolf2020transformers}. 
We first fine-tune our model on OCNLI~\cite{hu2020ocnli}, a Chinese natural language inference dataset before  fine-tuning on VGaokao, which has 8 epochs, with maximum input length 256, batch size 64, and learning rate 2e-5.

\section{Experiments}

We conduct experiments on our proposed VGaokao dataset and compare our Extract-Integrate-Compete approach with several baselines.

\textbf{RoBERTa-Large-Chunk} \cite{liu2019roberta} is an end-to-end method without explicit evidence retrieval. This model splits the long passages into fix-length chunks of 200 tokens. {Candidate answers are obtained from each chunk using existing MRC models, which are further aggregated over all chunks}. We use the pre-trained Chinese RoBERTa-wwm-ext-Large~\cite{cui2019pre}.

\textbf{BM25}~\cite{robertson2009probabilistic}
is a bag-of-words retrieval method, which uses sparse features to retrieve evidence sentences. We use the version implemented by Pyserini\footnote{\url{https://github.com/castorini/pyserini}}.

\textbf{Sent-BERT} ~\cite{reimers2019sentence} uses BERT to obtain contextualized dense representations for the texts and retrieve evidence sentences via cosine similarity.

\textbf{BeamDR}~\cite{xiong2021answering, zhao2021multi} 
is an iterative evidence selection technique with beam search and dense retrieval. It updates the query by appending the newly-extracted evidence in each iteration.

The \textbf{BM25}, \textbf{Sent-BERT}, and \textbf{BeamDR} are evidence extraction methods, which are combined with our proposed pair-wise option competition method to obtain the final question-level results.
For \textbf{BM25} and \textbf{Sent-BERT}, which cannot address the problem of multiple evidence, we report their experiment results with top 1 or top 2 evidence sentences selected.
To make fair comparison, we do not intensively tune any models, including ours. Specifically, we first search hyperparameters for the \textbf{BM25} model, and apply the same hyperparameters to other baselines and our method. 

We use three metrics to evaluate evidence quality on the subset annotated with golden evidence sentences: precision (\textbf{P}), recall (\textbf{R}), and F1 (\textbf{F1}). 
The accuracy of predicted answers (\textbf{Acc.}) is used to evaluate the performance on the question level.

\begin{table}
\centering
\small
\setlength{\tabcolsep}{5pt}
\begin{tabular}{l|ccc|c}
\toprule
&\multicolumn{3}{c|}{Evidence Metrics}& \multicolumn{1}{c}{QA Metrics}\\ \cmidrule{2-5}
& \textbf{P} & \textbf{R} & \textbf{F1} &  \textbf{Acc.}\\
\midrule
RoBERTa-L-Chunk & --- & --- & --- & 41.9\\
\midrule
BM25 Top 1 & \textbf{93.0} & 71.1 & 77.9  & 47.6 \\
BM25 Top 2 & 64.5 & \textbf{87.7} & 71.6  & 48.4 \\
Sent-BERT Top 1 & 87.0 & 67.3 & 73.4 & 48.4 \\
Sent-BERT Top 2 & 55.5 & 79.1 & 62.8 & 48.9 \\
\midrule
BeamDR & 52.5 & 74.0 & 59.2 & 48.9 \\ \midrule
Hard Masking & 82.5 & 79.5 & \textbf{79.2} & 49.3 \\
Soft Masking & 82.5 & 77.5 & 78.0 & \textbf{49.6} \\
\bottomrule
\end{tabular}

\caption{Performance of baselines and our methods on VGaokao (\%). RoBERTa-L-Chunk is short for RoBERTa-Large-Chunk. }
\label{tab:exp}
\end{table}

\subsection{Main Results}
\label{sec:main_results}

As we can see in Table~\ref{tab:exp}, our Extract-Integrate-Compete approach outperforms the end-to-end baseline and other evidence retrieval methods in both evidence quality and answer prediction accuracy. We analyze the results by comparing 1) passage chunking and evidence selection, 2) one-off evidence selection and iterative evidence selection, and 3) different query updating strategies.

\paragraph{Passage Chunking vs. Evidence Selection} 

RoBERTa-Large-Chunk simply chunks passages into pieces, which is a common practice for end-to-end MRC models~\cite{bert, kwiatkowski2019natural}. It obtains a question-level accuracy of 41.9\%, lagging behind the methods with additional evidence extraction step by at least 5.7\%.
We think it is because 
the evidence sentences of a statement may appear in different chunks. 
Without evidence selection, models have to split the long article, which results in 
incomplete evidence for training and harm the performance.
Even such simple methods as BM25 outperform the chunking method thanks to the retrieved evidence that is more relevant to the statement. 
Moreover, by introducing evidence selection, we can have a small-sized but focused supportive candidates to be feed the MRC model, greatly reducing required computational resources.
These results illustrate the necessity of selecting evidence pieces from the long articles before performing verification.

\begin{table*}[tb]
\centering
\small
\setlength{\tabcolsep}{5pt}
\begin{tabular}{p{0.98\textwidth}}

\toprule 
{\small \textbf{Statement:}} 
{\small \red{互联网颠覆传统制造业的现象在发达国家已出现；}\blue{国内互联网巨头主动涉足传统制造业，互联网去工业化初现端倪}。}

{\small \red{The phenomenon of the Internet subverting traditional manufacturing has already appeared in developed countries;} \blue{domestic Internet giants have taken the initiative to set foot in traditional manufacturing, and the de-industrialization of the Internet has begun to take shape}.} \\
{\small \textbf{S-BERT Top 1 Evidence:} ② } \quad\quad 
{\small \textbf{Soft Masking Evidence:} ①②}     \quad\quad
{\small \textbf{Golden Evidence:} ①②}\\
{\small \blue{\textbf{①}}\blue{互联网的去工具化从百度、腾讯等互联网巨头纷纷主动涉足传统制造业已经初现端倪。}}

{\small \blue{The de-instrumentation of the Internet has begun to emerge from Internet giants such as Baidu and Tencent, which take the initiative to set foot in traditional manufacturing.}} \\
{\small \red{\textbf{②}}\red{而互联网对传统制造业带来的颠覆在发达国家也已出现。}}

{\small \red{The disruption of the traditional manufacturing industry brought by the Internet has also appeared in developed countries.}} \\

\midrule
{\small \textbf{Statement:}} 
{\small \red{美国某款新能源汽车生产者运用了物联网概念，取消了4S店的商业模式，自己销售产品并提供保养、维修等各项服务。}}

{\small \red{A new-energy vehicle manufacturer in the United States used the concept of the Internet of Things, abandoned the business model of the 4S shop, sold its own products and provided various services such as maintenance and repair.}} \\
{\small \textbf{S-BERT Top 2 Evidence:} ①②}     \quad\quad
{\small \textbf{Soft Masking Evidence:} ①}     \quad\quad
{\small \textbf{Golden Evidence:} ①} \\
{\small \red{\textbf{①}}\red{美国的某款新能源汽车，由于运用了物联网概念，已经取消了传统的4S店商业模式，不仅销售不需要，甚至保养、维修也不再需要4S店。}} \\

{\small \red{A new energy vehicle in the United States, due to the use of the Internet of Things concept, has abandoned the traditional 4S shop business model. Not only does it not need to be sold, but even maintenance and repairs do not require 4S shops.}} \\
{\small \textbf{②}在工业4.0阶段，互联网已经不再是传统意义上的信息网络，更是物质、能量和信息互相交融的物联网，传递的也不仅是传统意义上的信息，还可以包括物质和能量的信息。}

{\small In the 4.0 stage of industry, the Internet is no longer an information network in the traditional sense. It is an Internet of Things that integrates material, energy and information. It transmits not only information in the traditional sense, but also  information about matter and energy.} \\

\midrule
{\small \textbf{Statement:}} 
{\small \red{佛教禅宗的坐忘、顿悟等与空灵美的琼澈晶莹境界看来相似，}\blue{实则相反，因二者赖以存在的基础完全不同。}}

{\small \red{Buddhism Zen's ``sit and forget'' and ``insight'' are similar to crystal artistic realm of ethereal beauty,} \blue{but
they are opposite, because the bases of their existence are completely different}.} \\
{\small \textbf{BeamDR Evidence:} ①②    } \quad\quad
{\small \textbf{Soft Masking Evidence:} ①③     } \quad\quad
{\small \textbf{Golden Evidence:} ①③    } \\
{\small \red{\textbf{①}}\red{佛教禅宗的“坐忘”“顿悟"与空灵之美的那种琼澈晶莹的艺术境界，从表面上看，颇有些异曲同工之处。}}

{\small \red{On the surface, there are some similarities between the ``sit and forget'' and ``insight'' of Zen Buddhism, and the crystal clear artistic realm of ethereal beauty.}} \\
{\small \textbf{②}空灵美不是老庄的“虚静”，佛教的“空忘”那样引人出世的“禅意”，而是一种实与虚、有限与无限的契合统一。}

{\small This kind of ethereal beauty is not the ``emptiness and quietness'' of Taoism or the ``emptiness and forgetfulness'' of Buddhism, but a combination of reality and emptiness, of limitation and infinity.} \\
{\small \blue{\textbf{③}}\blue{实际上，佛道两教是主观唯心主义的东西，空灵之美则恰恰相反，它根植于绚丽多彩的现实生活。}}

{\small \blue{In fact, Buddhism and Taoism are subjective idealism, while ethereal beauty is just the opposite and rooted in colorful real life.}}\\

\bottomrule
\end{tabular}
\caption{Case study of the evidence retrieved by different methods. {Golden evidence and its corresponding part in the statement are marked with the same color (red or blue).} }
\label{tab:case_study}
\end{table*}

\paragraph{One-off Selection vs. Iterative Selection} 
In Table~\ref{tab:exp}, our methods, with either hard masking or soft masking, outperform one-off evidence selection methods, BM25 and Sent-BERT settings,
by 2\textasciitilde16\% in the evidence extraction F1. 

Specifically, by selecting the top 1 evidence sentence, these methods ensure a high precision but fail to provide sufficient evidence to the statements that require more evidence sentences.
For example, in the first case of Table~\ref{tab:case_study}, the top 1 evidence sentence selected by S-BERT only covers the second half of the statement while our method with soft masking strategy succeeds in retrieving the complete evidence chain with iterative extraction.

On the other hand, by selecting top 2 evidence sentences, Sent-BERT and BM25 achieve a high recall but may introduce irrelevant evidence to the statements requiring fewer evidence sentences. 
For the second case in Table~\ref{tab:case_study}, using the top 2 evidence sentences will introduce a noisy sentence,
while our method correctly determines that the statement only needs one evidence sentence with the adaptive evidence integrator.

Our evidence selection module achieves a balance between the precision and recall of the retrieved evidence pieces, 
obtaining the highest F1 scores among all methods. 
By iteratively extracting possible partial evidence, our methods ensure a relatively high recall of over 80\%; by adaptively integrating truly complementary evidence pieces, our methods lead to a relatively high precision of nearly 80\%. 
Therefore, our methods with iterative extraction and adaptive integration achieve the highest F1 scores among competitive counterparts. 
Besides, the higher evidence performance also results in better QA performance, with an improvement of more than 0.7\%.
This indicates that carefully retrieved evidence produces more reliable evidence chains for the verification module, improving the explainability of our methods.

\paragraph{Query Updating Strategy} 
Our query updating strategies, hard masking and soft masking, outperform BeamDR by approximately 20\% in evidence retrieval F1. After each iteration, BeamDR appends the extracted evidence sentence to the query 
for following iterations.
This strategy may put more emphasis on the overlap between retrieved evidence and queries, while the unmatched parts are likely to receive less attention.
This tendency prevents the model from gathering complementary evidence sentences.
By contrast, our query updating methods highlight the tokens in the query whose corresponding evidence is still missing, thus, being more likely to retrieve complementary evidence pieces. 

For the third case of Table~\ref{tab:case_study}, the second evidence sentence returned by BeamDR shares the same topic with the first one, but irrelevant to the given statement. 
On the contrary, the second evidence sentence returned by our method is a complementary one that supports the second half of the statement. A possible reason is that our soft masking strategy can 
highlight the phrases such as \textit{opposite}, \textit{basis of existence} that are absent in the first retrieved evidence, pushing the retriever to collect complementary 
evidence.

Our two masking strategies, hard masking and soft masking, perform comparably on both metrics.
The hard masking is easy to implement and requires little computation when updating queries; the soft masking can to some extent maintain the information in the matched tokens, which may be useful in the next step evidence retrieval. To summarize, our query updating strategies can make the retrieval target of each iteration more focused and boost the performance in both evidence extraction and the final answer selection.

\subsection{Ablation Study}
To reveal the effectiveness of each module in our approach, we conduct experiments with three ablated settings with soft masking for the iterative extractor.
As we can see in Table~\ref{tab:ablation},
After we remove each of the three modules, the performance drops substantially in evidence F1 and question-level accuracy, demonstrating the effectiveness of each part in our approach. 

Specifically, after removing the iterative extraction module, we can observe a decrease of 6.3\% in the evidence F1 and a decrease of 1.9\% in the question-level accuracy, because one-off evidence extraction fails to gather sufficient complementary evidence pieces. Thus the option competition module is unlikely to make a thoughtful decision based on an incomplete evidence chain. 

When we remove the adaptive integration module, the evidence F1 drops by 15.1\% and the question-level test accuracy 
drops by 2.2\%. This may result from irrelevant evidence pieces introduced in the later iterations of evidence extraction. The adaptive integration module can screen out this kind of noisy information by evaluating the associativity between extracted evidence, i.e., the property that evidence pieces are logically connected to cover the information in the query.

Without the pairwise option competition, the question-level accuracy decreases by 0.9\%. This indicates that pairwise option competition is superior to independent treatment of each option, because pairwise competition can push the model to capture the subtle difference between options.

\begin{table}
\centering
\small
\setlength{\tabcolsep}{9pt}
\begin{tabular}{l|l|l}
\toprule
\textbf{Methods} &  \textbf{Evi. F1}  & \textbf{Qu. Acc.}\\
\midrule
Soft Masking &  78.0 &  49.6 \\
\midrule
w/o Iterative Extraction  & 71.7 {\scriptsize (~~-6.3)}  &  47.7 {\scriptsize (-1.9)}\\
w/o Adaptive Integration  &  62.9 {\scriptsize (-15.1)} &   47.4 {\scriptsize (-2.2)} \\
w/o Pairwise Competition &  --- & 48.7 {\scriptsize (-0.9)} \\
\bottomrule
\end{tabular}
\caption{Ablation study 
of our soft masking method on VGaokao. 
}
\label{tab:ablation}
\end{table}

\section{Related Works}
Several previous datasets are constructed from different subjects of Gaokao, including Geography~\cite{huang2019geosqa}, English~\cite{lai2017race, sun2019dream}, and History~\cite{guo2017effective}. 
Different from these efforts, we focus on Chinese subject to introduce native-speaker level reading comprehension challenges.

Evidence extraction based methods have been studied in MRC~\cite{talmor2018web, perez2020unsupervised} and fact verification~\cite{thorne2018fever}.
In this work, we introduce evidence extraction to our method for solving the long article challenge in VGaokao.

Iterative evidence extraction can be seen as a sort of question decomposition method, a technique widely used in QA tasks with complex questions
\cite{talmor2018web, perez2020unsupervised}.
However, in VGaokao, the queries may interweave by implicit semantic relationship, so that models could not explicitly separate the queries into independent sub-queries.
We thus adopt an iteative extractor with an adaptive integrator to decompose queries in an implicit way.

Another stream of works adopt an iterative framework by updating the queries. \citet{qi2019answering} iteratively generate new queries by selecting a span from the question and retrieved evidence while \citet{xu2019enhancing,xiong2021answering, zhao2021multi} directly append retrieved evidence to the query.
Compared with these works,
we introduce two novel query updating techniques, hard masking and soft masking, together with an evidence integration module to avoid too much 
overlap between evidence and to dynamically determine the number of required evidence sentences.

\section{Conclusion}
In this paper, we present a novel verification style reading  comprehension  dataset named VGaokao from the Chinese Language tests of Gaokao for Chinese native speakers, which embed multiple advanced language understanding skills. 
To address the challenges in VGaokao, we propose a new Extract-Integrate-Compare approach for complementary evidence retrieval/integration and option discrimination. Experiments show that our approach outperforms several strong baselines, with additional merits of efficiency and explainability. 
We believe VGaokao is a challenging test-bed for natural language understanding in Chinese and {encourage further research in verification style reading comprehensionn}.

\section*{Acknowledgements}
We thank the anonymous reviewers for the helpful comments and suggestions. 
This work is supported by Hi-Tech R\&D Program of China (No.2018YFB1005100).

\bibliography{emnlp2021}
\bibliographystyle{acl_natbib}

\end{CJK}
\end{document}